\documentclass[conference]{IEEEtran}
\IEEEoverridecommandlockouts

\usepackage{cite}
\usepackage{amsmath,amssymb,amsfonts}
\usepackage{algorithmic}
\usepackage{algorithm}
\usepackage{array}
\usepackage{graphicx}
\usepackage{textcomp}
\usepackage{xcolor}
\usepackage{multirow}

\def\BibTeX{{\rm B\kern-.05em{\sc i\kern-.025em b}\kern-.08em
    T\kern-.1667em\lower.7ex\hbox{E}\kern-.125emX}}
\begin{document}

\title{Efficient RF Passive Components Modeling with Bayesian Online Learning and Uncertainty Aware Sampling\thanks{This work was supported by the Natural Science Foundation of China (Grant No. 62074100).}}
\author{
Huifan Zhang, Pingqiang Zhou\\
ShanghaiTech University, Shanghai, China \\
\{zhanghf, zhoupq\}@shanghaitech.edu.cn
}
\maketitle
\begin{abstract}
Conventional radio frequency (RF) passive components modeling based on machine learning requires extensive electromagnetic (EM) simulations to cover geometric and frequency design spaces, creating computational bottlenecks. In this paper, we introduce an uncertainty-aware Bayesian online learning framework for efficient parametric modeling of RF passive components, which includes: 1) a Bayesian neural network with reconfigurable heads for joint geometric-frequency domain modeling while quantifying uncertainty; 2) an adaptive sampling strategy that simultaneously optimizes training data sampling across geometric parameters and frequency domain using uncertainty guidance. Validated on three RF passive components, the framework achieves accurate modeling while using only 2.86\% EM simulation time compared to traditional ML-based flow, achieving a 35$\times$ speedup.
\end{abstract}


\section{Introduction}

Radio frequency integrated circuits (RFICs) form the cornerstone of modern communication systems, enabling critical technologies from 5G/6G networks to Internet-of-Things (IoT) devices \cite{chae2025ml}. As operational frequencies increase into millimeter-wave and terahertz regimes, traditional lumped-element circuit models become inadequate in mm-wave circuits. Consequently, designers must resort to computationally intensive full-wave electromagnetic (EM) simulations to ensure the accurate modeling of RF passive components.

Conventional RF system optimization relies heavily on iterative manual tuning which is a time-consuming process requiring significant domain expertise. This contrasts obviously with digital circuit design, which benefits from mature automated synthesis tools. Recent efforts have explored parametric modeling of microwave structures using machine learning (ML) techniques \cite{shibata2022novel, feng2015parametric, kim2024traceformer}. Neural networks can model EM behaviors, e.g. Scattering (S-) parameters, taking geometric variables and frequency as input features. In PulseRF \cite{chae2024pulserf}, Hyusnu et al. employ a physics-augmented UNet-based ML surrogate model to replace EM simulators, the models are trained with a dataset of 4,000 and each simulation for a 6-port network consumes up to 20 minutes. In the work by Ren et al. which utilizes a novel convolutional-autoencoder ML model to calculate, the dataset of modeling one structure contains over 6,000 cases \cite{shibata2022novel}. These approaches still require extensive training datasets derived from costly EM simulations, diminishing their practical utility for covering multi-dimensional design spaces. To address this issue, Qipan et al. employ cross-validation ensemble training to select a small set of representative samples to maintain high analysis accuracy \cite{wang2023mtl}. However, this approach requires redundant training across multiple data folds and only works on geometric domain.

\begin{figure}[tbp]
\centerline{\includegraphics[scale=0.65]{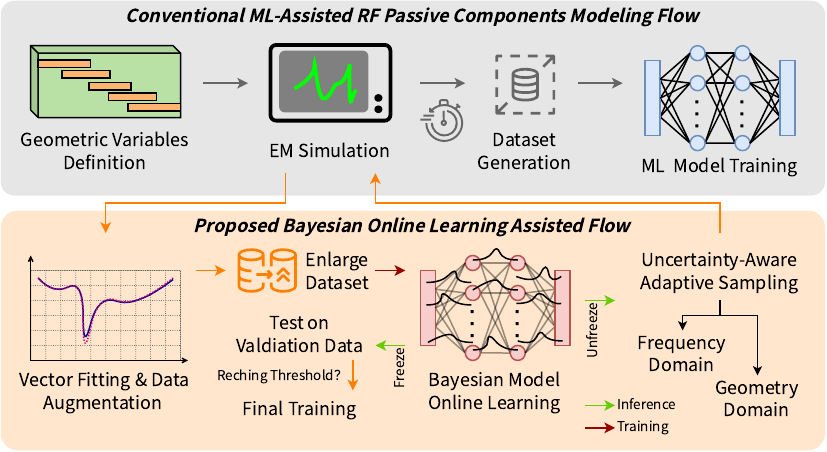}}
\caption{An illustration of conventional ML-assisted RF passive components modeling flow and proposed Bayesian online learning flow.}
\label{fig:Flow}
\end{figure}

Once the geometry is defined, full-wave EM simulations are performed to generate the frequency response of the EM behavior. These simulations demand significant computational resources, especially when covering a broad frequency spectrum. Vector fitting \cite{gustavsen1999rational, lei2010decade, ravula2023one} is a numerical technique that reconstructs a frequency response by interpolating the responses at specific frequency points. The selection of sampling frequencies presents a key challenge in vector fitting: too few sample points lead to inaccurate approximations, while too many negate the computational efficiency benefits. To address this challenge, adaptive frequency sampling (AFS) methods are employed to identify a minimal yet sufficient set of frequency points. Building on the foundational work by Dhaene et al. \cite{dhaene1995adaptive}, later improvements include local rational modeling with efficient uncertainty estimation \cite{peumans2024adaptive} and and Bayesian linear regression to replace traditional least squares in vector fitting \cite{de2019bayesian}. Here, uncertainty quantifies the degree of confidence or potential deviation between the fitted model and the true EM response.

Despite these advances in frequency-domain adaptive sampling, two critical limitations remain: 1) the AFS method requires iterative vector fitting loops to determine optimal sampling frequencies and 2) existing uncertainty estimation methods focus solely on frequency response while neglecting dependencies on geometric and material parameters. Consequently, adaptive sampling that simultaneously considers both geometric parametric and frequency domains remains underexplored. 

Bayesian neural networks (BNNs) \cite{blundell2015weight, depeweg2018decomposition} offer a promising method for uncertainty quantification in deep learning. BNNs provide robust uncertainty quantification by estimating output distributions rather than point estimations, enabling their successful applications in computer vision \cite{kendall2017uncertainties}, civil engineering \cite{bateni2007bayesian}, and also in electronics \cite{aminian2001fault}. Yushi et al. \cite{liu2022efficient} employ a BNN as an alternative to Gaussian process for antenna design optimization. However, their approach focuses only on modeling limited performance metrics, such as the maximum reflection coefficient or realized gain, and does not account for the full parametric frequency response.

This paper presents a novel BNN approach for modeling S-parameter responses of RF passive components. We develop an online learning framework that iteratively selects optimal geometric parameters and frequency points for EM simulation, thereby constructing a minimal yet sufficient training dataset. Our core contributions include:

\begin{itemize}
    \item We propose a reconfigurable BNN architecture to model the parametric S-parameters frequency response of different RF components. To minimize training overhead, we implement an online learning framework that iteratively optimizes stochastic weights and biases.
    \item Based on the BNN uncertainty estimates,\textit{ for the first time,} we develop an uncertainty-aware adaptive sampling method that adaptively selects training points across both geometric and frequency domains, using only 2.86\% of EM simulation time compared to conventional ML-based modeling method.
    \item We demonstrate the accuracy of our proposed modeling methodology on three test structures: bandpass coupled line filter (BCLF), spiral inductor (SI), and microstrip transmission lines (MTL). Our method performs accurate modeling compared to the conventional ML method, (e.g., 32\% RMSE reduction for SI and 1.9\% for BCLF).
\end{itemize}

\section{Backgrounds}
\subsection{Vector Fitting}

Vector fitting \cite{gustavsen1999rational, lei2010decade} is a widely used technique for broadband modeling of frequency-domain responses in RF systems, enabling accurate characterization of critical components such as filters, antenna arrays and frequency converters. The algorithm approximates the frequency response by a complex-valued rational function of the form
\begin{equation}
    h(s) = \boldsymbol{d} + s\boldsymbol{e} + \sum_{n=1}^{N}\frac{\boldsymbol{r}_n}{s-\boldsymbol{p}_n}
    \label{vf_base}
\end{equation}
where $h(s):\mathbb{C}\to\mathbb{C}^{p\times p}$ represents the frequency response of different port combinations in Laplace domain, $\boldsymbol{d}$ is the DC offset, proportional term $\boldsymbol{e}$ is the linear term, poles $\boldsymbol{p}_n$ and residues $\boldsymbol{r}_n$ are real or come in conjugate pairs. Starting from initial pole estimates, all coefficients are iteratively solved via least-squares optimization \cite{gustavsen1999rational}. Algorithmic improvements include accelerated pole relocation for faster convergence \cite{gustavsen2006improving} and parallelization strategies for large-scale systems \cite{chinea2011parallelization}. 

\subsection{Adaptive Frequency Sampling}
Except from algorithmic refinements mentioned above, frequency sample selection critically improves fitting accuracy while reducing the sampling number. AFS addresses this by dynamically concentrating samples in regions of high response variability through uncertainty quantification \cite{dhaene1995adaptive}. Existing AFS methods \cite{antonini2008broadband} typically implement the following procedure:
\begin{enumerate}
\item Construct an ensemble of rational approximations $\{ h_k(s) \}_{k=1}^K$ with varying orders $N_k$ where $k$ is the number of the model.
\item Quantify frequency-dependent uncertainty $\mathcal{U}(s)$ by computing the variance of the ensemble at each angular frequency $s$. A common measure is the maximum root mean square deviation (RMSD) across different models:
\begin{equation}
    \mathcal{U}(s) = \max_{m, n} \sqrt{\frac{1}{n_s} \sum_{s} \| h_m(s) - h_n(s) \|_F^2},
    \label{eq:uncertainty}
\end{equation} 
where $n_s$ represents the number of sampling frequencies and $\|\cdot\|_F$ denotes the Frobenius norm for multi-port systems.
\item  Iteratively refine the sampling density in frequency regions where $\mathcal{U}(s)$ exceeds a threshold, allocating additional samples proportional to the local uncertainty.
\end{enumerate}

The nested AFS method, which requires $\mathcal{O}(N_{max}^2\cdot n_s)$ iterations in total, becomes prohibitively expensive for high-order models or wide frequency ranges. Therefore, a \textit{one-shot} uncertainty estimation method that computes the uncertainty without repetitive vector fitting invocations is required. 

\subsection{Bayesian Neural Networks}
BNNs \cite{goan2020bayesian, jospin2022hands} are stochastic neural networks trained using a Bayesian approach. Given a set of training examples $\mathcal{D} = (\boldsymbol{x}_i, \boldsymbol{y}_i)$, conventional artificial neural networks (ANNs) employ deterministic weight parameters $\mathbf{w}^*$ optimized through maximum likelihood estimation:
\begin{equation}
\mathbf{w}^* = \arg \max_{\mathbf{w}} \log P(\mathcal{D}|\mathbf{w})   
\end{equation}
yielding point estimations $f(\boldsymbol{x}; \mathbf{w}^*)$ without inherent uncertainty measures. However, BNNs replace the model parameters (weights, biases, etc.) with random variables. The corresponding posterior distribution $p(\mathbf{w}|\mathcal{D})$ are computed by applying Bayes' theorem:
\begin{equation}
p(\mathbf{w}|\mathcal{D})=\frac{p(\mathcal{D}_{\boldsymbol{y}}|\mathcal{D}_{\boldsymbol{x}},\mathbf{w})p(\mathbf{w})}{\int_{\mathbf{w}} p(\mathcal{D}_{\boldsymbol{y}}|\mathcal{D}_{\boldsymbol{x}},\mathbf{w'}) p(\mathbf{w'}) \, d\mathbf{w'}}    
\end{equation}
where the dataset inputs and outputs are denoted as $\mathcal{D}_{\boldsymbol{x}}$ and $\mathcal{D}_{\boldsymbol{y}}$, respectively. Unfortunately, the evidence $\int_{\mathbf{w}} p(\mathcal{D}_{\boldsymbol{y}}|\mathcal{D}_{\boldsymbol{x}},\mathbf{w'}) p(\mathbf{w'}) \, d\mathbf{w'}$ is difficult to be calculated analytically, or even efficiently sampled from \cite{graves2011practical}. Variational inference addresses this problem by finding the approximate distribution. First, a simple parameterized distribution $q_\theta(\mathbf{w})$ is selected. Then, the Kullback-Leibler (KL) divergence which measures their closeness is minimized \cite{blei2017variational}: 
\begin{align}
    \theta^* &= \arg \min _{\theta} \text{KL}\big( q_\theta(\mathbf{w}) \parallel p(\mathbf{w}|\mathcal{D}) \big) \\
    &= \arg \min _{\theta}  \mathbb{E}[q_\theta(\mathbf{w})]-\mathbb{E}[\log p(\mathbf{w},\mathcal{D})]+\log p(\mathcal{D}).
\end{align}
which is equivalent to maximizing the evidence lower bound (ELBO):
\begin{equation}
\text{ELBO}(\theta) =\mathbb{E}_{q_\theta(\mathbf{w})} \big[ \log p(\mathcal{D}|\mathbf{w}) \big] - \text{KL}\big( q_\theta(\mathbf{w}) \parallel p(\mathbf{w}) \big)
\label{eq:elbo}
\end{equation}

\section{Methods}
This section details our proposed methodology. Section III.A presents the key concepts of our workflow and highlights its conceptual differences from previous ML-based modeling approaches. Section III.B introduces the Bayesian neural network architecture and the corresponding online learning strategy. Section III.C elaborates on the implementation of the uncertainty-aware sampling method.

\subsection{Overall Flow}
\begin{figure}[htbp]
\centerline{\includegraphics[width=\linewidth]{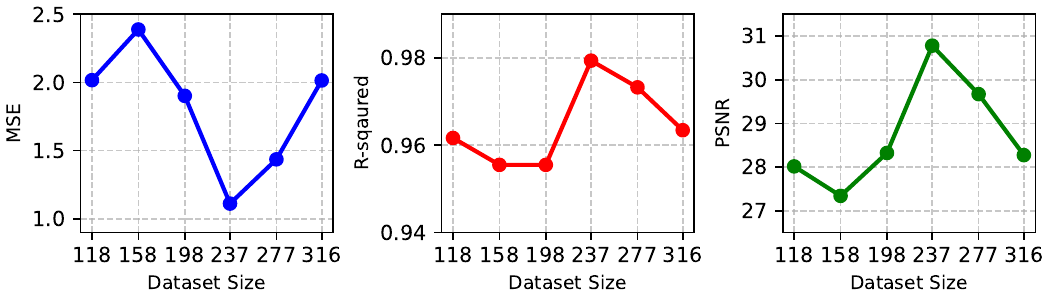}}
\caption{Relationship between test accuracy (evaluated using MSE, R-squared, and PSNR) and training dataset size for a spiral inductor.}
\label{fig:Accuracy}
\end{figure}

ML-based RF components modeling methods aim to replace computationally expensive EM simulations with rapid evaluations. However, training such models still requires extensive labeled datasets generated by experiments or EM simulators (e.g., Ansys HFSS). Furthermore, expanding the training dataset does not consistently improve the accuracy of the model. As shown in Fig.~\ref{fig:Accuracy}, when modeling a spiral inductor, the test accuracy increases as the dataset grows but declines with further expansion due to overfitting. 

To address these limitations, we propose a Bayesian online learning framework with uncertainty-aware sampling, as illustrated in Fig. \ref{fig:Flow}. Our methodology begins by establishing the geometric and frequency parameter space, followed by generating initial training and validation datasets through full-wave EM simulations. Our proposed \textit{online learning} process then iteratively executes three steps: 1) training the BNN to learn the relationship between input features and responses of the S parameter, 2) quantifying uncertainty in unexplored regions of the design space, and 3) adaptively selecting the next batch of simulation points using our uncertainty-aware adaptive sampling method. This training and sampling loop continues until either the validation error falls below a predetermined threshold or the maximum iteration count is reached, at which point we perform final training on the accumulated dataset to produce the ultimate surrogate model. For computational efficiency, only the most recent batch of data is used to train the BNN in each iteration.

\subsection{Proposed Bayesian Neural Network Structure}

The primary function of our BNN is is to establish a probabilistic mapping between input parameters (geometric dimensions and frequency) and the corresponding S-parameter responses. There are typically two approaches to modeling the frequency response of RF systems: 
\begin{enumerate}
    \item \textbf{Point} mode: Treating frequency as an input feature alongside material/geometric parameters;
    \item \textbf{Vector} mode: Encoding the full frequency response as vectorized representations.
\end{enumerate}
For the first approach, fully-connected neural networks (FCNNs) excel at learning global feature interactions in low-dimensional data \cite{wang2023mtl}. For the second approach, convolutional architectures effectively handle structured spectral data, as demonstrated by transposed CNNs in \cite{torun2019spectral} and U-Net variations in \cite{chae2024pulserf, shibata2022novel}. 
\begin{figure}[htbp]
\centerline{\includegraphics[scale=0.65]{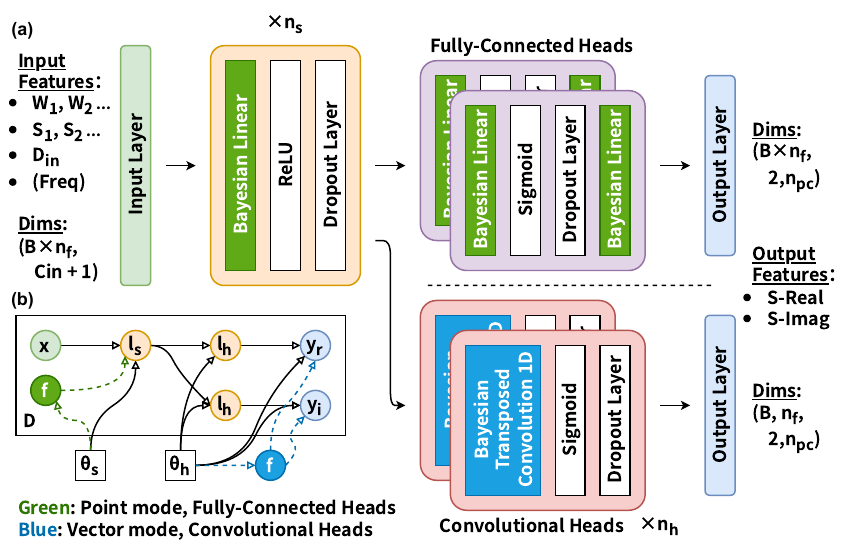}}
\caption{(a) Our proposed Bayesian neural network (BNN) architecture with reconfigurable heads, including fully-connected heads for point mode and convolutional heads for vector mode. (b) Probabilistic graphical model of the proposed BNN structure, where green and blue components represent the fully-connected and convolutional head configurations, respectively.}
\label{fig:Network}
\end{figure}

\begin{figure*}[tbp]
\centerline{\includegraphics[width=\textwidth]{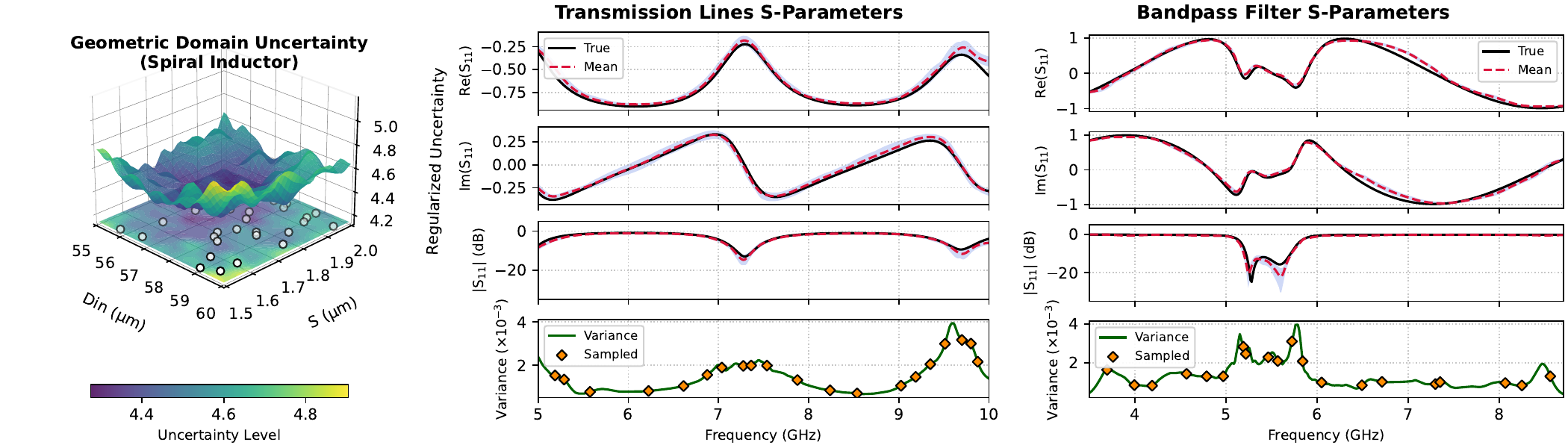}}
\caption{(1) Estimated uncertainty distribution across the geometric domain for the spiral inductor test case. (2) BNN inference results showing: (i) sampled transmission line and bandpass coupled-line filter responses, and (ii) their corresponding sampling frequencies. The blue shaded region indicates the min-max range of predicted S-parameters, while the dashed line represents the mean response across all samples.}
\label{fig:Uncertainty}
\end{figure*}

As shown in Fig.~\ref{fig:Network}(a), we propose a Bayesian neural network with a shared backbone and \textit{reconfigurable} heads to handle both modeling paradigms. The backbone comprises stacked Bayesian linear blocks. Unlike traditional linear layers with deterministic parameters:
\begin{equation}
\boldsymbol{y} = \mathbf{W}\boldsymbol{x} + \mathbf{b}
\end{equation}
where the weight matrix $\mathbf{W} \in \mathbb{R}^{d_o \times d_i}$ and bias $\mathbf{b} \in \mathbb{R}^{d_o}$ are learnable tensors, our Bayesian formulation treats weights and biases as stochastic random variables:
\begin{equation}
\mathbf{W} \sim \mathcal{N}(\boldsymbol{\mu}_W, \boldsymbol{\sigma}_W^2), \quad \mathbf{b} \sim \mathcal{N}(\boldsymbol{\mu}_b, \boldsymbol{\sigma}_b^2).
\end{equation}
The normal distribution $\mathcal{N}(\boldsymbol{\mu}, \boldsymbol{\sigma}^2)$ is commonly adopted for its mathematical tractability, although other distributions can be substituted. Given the prior distribution parameter $\boldsymbol{\sigma}_p$, $\boldsymbol{\mu}_p$, and learning rate $\alpha$, the gradients are derived using Bayes-by-backprop: 
\begin{align}
f(\theta) & = \log p(\mathcal{D}|\theta) - \text{KL}\big( \mathcal{N}(\theta|\mathcal{D}) \parallel p(\theta) \big) \\
& = \log p(\mathcal{D}|\theta) - \big[\log\frac{\boldsymbol{\sigma}}{{\boldsymbol{\sigma}_p}} + \frac{\boldsymbol{\sigma}_p^2 + (\boldsymbol{\mu} - \boldsymbol{\mu}_p)^2}{2\boldsymbol{\sigma}_p^2} - \frac{1}{2} \big], \\
\Delta_\theta f & = \text{backpropagate}_\theta(f),\\
\theta & = \theta - \alpha \Delta_\theta f.
\end{align}

With regard to the transposed convolution layer, the method is similar that every items in the convolutional kernel $\mathbf{K}\in \mathbb{R}^{c_i \times c_o \times k \times k}$ and vector bias $\mathbf{b} \in \mathbb{R}^{c_o}$ are transformed into independent random variables.

\subsection{Uncertainty Aware Sampling}

Conventional adaptive sampling approaches for global surrogate modeling typically utilize predicted uncertainty to sequentially generate new sample points. This paper introduces an adaptive sampling method that simultaneously 1) selects optimal geometric parameter configurations and 2) determines the frequency-specific simulation points required for EM simulation of each configuration. Our approach first generates an ensemble of $M$ BNN candidate ${f(s,\boldsymbol{x};\theta_m)}_{m=1}^M$ by sampling weights and biases from the posterior distribution $p(\theta|\mathcal{D})$. The predictive uncertainty is then quantified as:
\begin{equation}
\mathcal{U}(s, \boldsymbol{x}) = \sqrt{\frac{1}{M}\sum_{m=1}^M\|f(s,\boldsymbol{x};\theta_m) - \mu(s,\boldsymbol{x})\|_F^2}
\end{equation}
where $\mu(s,\boldsymbol{x}) = \frac{1}{M}\sum_{m=1}^M f(s,\boldsymbol{x};\theta_m)$ represents the ensemble mean prediction, $s$ denotes geometric parameters, $\boldsymbol{x}$ represents frequency parameters.

For geometric domain sampling, we aggregate uncertainties by summing across the entire spectrum: $\mathcal{U}(\boldsymbol{x}) = \sum_{s \in \mathcal{S}} \mathcal{U}(s, \boldsymbol{x})$, where $\boldsymbol{x}$ represents geometric parameters and $\mathcal{S}$ denotes the frequency domain. As visualized in Fig.~\ref{fig:Uncertainty}, we sample parametric configurations using an uncertainty-aware mixture distribution defined by the probability distribution function:
\begin{equation}
p(\boldsymbol{x_i}) = \lambda \cdot \frac{1}{N_d}+ (1 - \lambda) \cdot {\frac{\mathcal{U}(\boldsymbol{x_i})}{\sum_{j=1}^{N_d} \mathcal{U}(\boldsymbol{x_j})}}
\end{equation}
where $N_d$ is the number of elements in whole design space $\mathcal{D}$, and $\lambda \in [0,1]$ is a mixing parameter that balances uniform sampling and uncertainty-directed sampling.

For frequency spectrum sampling, our approach \textit{eliminates} the need for iterative vector fitting across multiple model orders required by conventional AFS methods. Leveraging the BNN model and GPU-accelerated parallel computing, we rapidly estimate uncertainty across the entire frequency spectrum in a single pass. As demonstrated in Algorithm~\ref{alg:afs} and visualized in Fig.~\ref{fig:Uncertainty}, our uncertainty-aware AFS algorithm first partitions the frequency domain based on cumulative uncertainty density and then performs random sampling within each partition to avoid redundant close sampling points.
\begin{algorithm}
\caption{Uncertainty-Aware AFS}
\label{alg:afs}
\begin{algorithmic}[1]
\REQUIRE $\boldsymbol{s}$: Laplace frequency range to sample, $\mathcal{U}(s)$: uncertainty values, $n_{\text{samples}}$: required sampling number
\STATE $\boldsymbol{S} \leftarrow [~]$ \COMMENT{Initialize sample array}
\STATE $\boldsymbol{C} \leftarrow \text{cumtrapz}(\mathcal{U}(s), \boldsymbol{s})$ \COMMENT{Cumulative uncertainty}
\STATE $\delta \leftarrow \boldsymbol{C}[-1]/n$ \COMMENT{Uncertainty per partition} 
\STATE $\boldsymbol{F}_{inv} \leftarrow \text{interpolate}(\boldsymbol{C}, \boldsymbol{s})$ \COMMENT{Inverse CDF}
\FOR {$i \leftarrow 0$ \TO $n_{\text{samples}} - 1$} 
\STATE $l \leftarrow \left\lfloor \boldsymbol{F}_{inv}(i \times \delta) \right\rfloor$ \COMMENT{Define partition range}
\STATE $r \leftarrow \left\lfloor \boldsymbol{F}_{inv}((i+1)\times \delta) \right\rfloor$ 
\STATE $\boldsymbol{S_i} \leftarrow \text{random}(l:r)$ \COMMENT{Randomly sample within the range}
\ENDFOR
\RETURN $\boldsymbol{S}$
\end{algorithmic} 
\end{algorithm}

Fig.~\ref{fig:Uncertainty} illustrates the normalized uncertainty distribution across both \textit{geometric} and \textit{frequency} domains, along with S-parameter estimations. In the first example, all parameters are fixed except coil inner diameter ($\text{D}_{\text{in}}$) and lap spacing (S). The uncertainty surface map is generated via BNN evaluation across the orthogonal design space. In the next two examples, the validation results on two representative EM structures are shown: bandpass filter circuits and transmission lines. The round and diamond markers indicate sampling locations selected by our uncertainty-aware AFS algorithm.


\section{Experiment Results}

\begin{figure}[htbp]
\centerline{\includegraphics[width=0.5\textwidth]{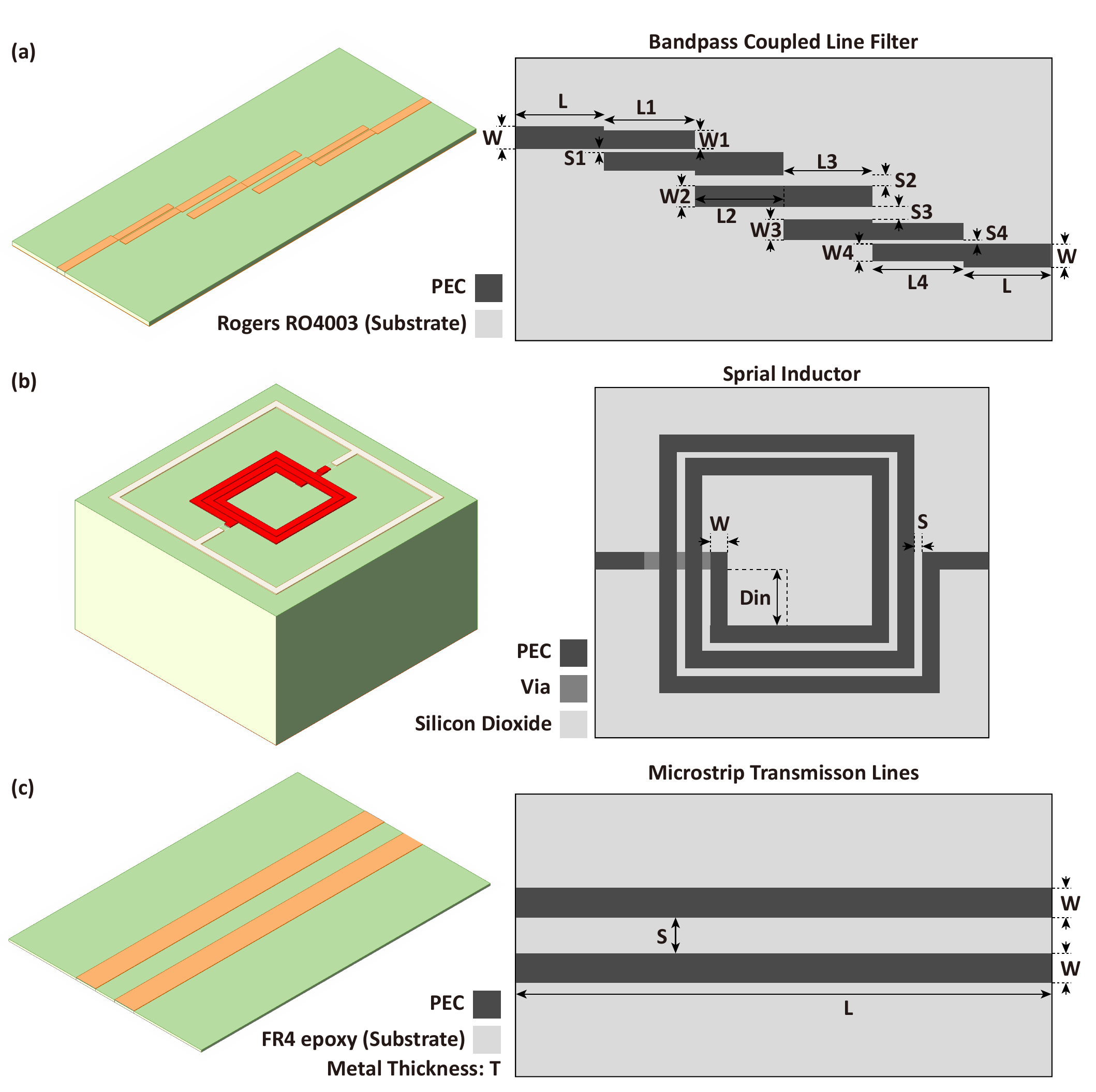}}
\caption{Three-dimensional models and top-view layouts of the experimental test structures: (1) bandpass coupled-line filter, (2) spiral inductor, and (3) microstrip transmission line.}
\label{fig:Structure}
\end{figure}

To demonstrate the generality of our approach across diverse RF passive components, we select three test cases: a \textit{bandpass coupled line filter} (BCLF), a \textit{spiral inductor} (SI), and a \textit{microstrip transmission line} (MTL) structure. Fig.~\ref{fig:Structure} shows their three-dimensional models and top-view layouts, while Table~\ref{tab:Parameters} lists their geometric parameter ranges. In the following sections, we first assess the impact of our proposed uncertainty-aware adaptive frequency sampling (UAW-AFS) method. Then we compare the final estimation accuracy and total EM simulations required to build the training dataset of our Bayesian online learning-assisted flow to a conventional ML-based flow. In this study, we employ simulation results generated by the commercial software Ansys HFSS as ground truth. All FC-NNs, CNNs, and BNNs are implemented on a computational cluster equipped with four Intel Xeon E5-269 CPUs and four NVIDIA GeForce GTX 1080 GPUs.
\begin{table}[htbp]
\caption{Parameters range of the experiment cases}
\begin{center}
\begin{tabular}{l l l c c c}
\hline\hline 
\textbf{Case} & \textbf{Symbol} & \textbf{Description} & \textbf{Min} & \textbf{Max} & \textbf{Step} \\
\hline
\multirow{8}{*}{\textbf{BCLF}} & L & \multirow{3}{*}{Line Length} & 7.7 mm & 7.9 mm & 0.1 mm\\
& L1, L4 & & 8.0 mm & 8.2 mm & 0.1 mm \\
& L2, L3 & & 7.8 mm & 8.0 mm & 0.1 mm  \\ 
& W & \multirow{3}{*}{Line Width} & 0.9 mm & 1.1 mm & 0.1 mm \\
& W1, W4 & & 0.7 mm & 0.9 mm & 0.1 mm \\
& W2, W3 & & 0.9 mm & 1.1 mm & 0.1 mm \\ 
& S1, S4 & \multirow{2}{*}{Line Distance} & 0.13 mm & 0.16 mm & 0.03 mm\\
& S2, S3 & & 0.55 mm & 0.6 mm & 0.05 mm\\
\hline
\multirow{3}{*}{\textbf{SI}} & Din & Inner Diameter & 55 um & 60 um & 1 um \\
& S & Lap Spacing & 1.5 um & 2.0 um & 0.1 um \\
& W & Lap Width & 15 um & 20 mm & 0.5 um \\
\hline
\multirow{4}{*}{\textbf{MTL}} & S & Line Distance & 1.5 mm & 2.0 mm & 0.1 mm \\
& L & Line Length & 25 mm & 30 mm & 1 mm \\
& T & Metal Thickness & 25 um & 30 um & 1 um \\
& W & Line Width & 2.5 mm & 3 mm & 0.1 mm \\
\hline
\hline
\end{tabular}
\end{center}
\label{tab:Parameters}
\end{table}
\subsection{Evaluation on proposed AFS Method}
In this experiment, metrics including mean absolute error (MAE), root mean square error (RMSE), and peak signal-to-noise ratio (PSNR), are used to evaluate the effectiveness of our proposed AFS method. We evaluate the fitting error of uniform sampling versus our proposed UAW method. As demonstrated in Table~\ref{tab:Afs}, the UAW-AFS approach consistently outperforms uniform sampling, achieving superior fitting accuracy in all test cases. 
\begin{table}[htbp]
\begin{center}
\caption{Error metrics comparison of vector fitting using uniform sampling and our proposed uncertainty-aware AFS.}
\label{tab:Afs}
\begin{tabular}{l l c c c}
\hline
\hline
\textbf{Case} & \textbf{Method} & \textbf{MAE} & \textbf{RMSE} & \textbf{PSNR} \\
\hline
\multirow{2}{*}{\textbf{BCLF}} 
    & Uniform         & $3.0248 \times 10^{-3}$ & $6.0809 \times 10^{-3}$ & $124.11$ \\
    & UAW             & $\mathbf{2.8133 \times 10^{-6}}$ & $\mathbf{5.7817 \times 10^{-6}}$ & $\mathbf{143.54}$ \\
\hline
\multirow{2}{*}{\textbf{SI}} 
    & Uniform         & $6.4515 \times 10^{-5}$ & $1.9140 \times 10^{-4}$ & $\mathbf{129.48}$ \\
    & UAW             & $\mathbf{1.4065 \times 10^{-5}}$ & $\mathbf{4.0047 \times 10^{-5}}$ & $127.90$ \\
\hline
\multirow{2}{*}{\textbf{MTL}} 
    & Uniform         & $2.7759 \times 10^{-2}$ & $1.1417 \times 10^{-1}$ & $70.053$ \\
    & UAW             & $\mathbf{3.8992 \times 10^{-4}}$ & $\mathbf{1.5019 \times 10^{-3}}$ & $\mathbf{73.275}$ \\
\hline
\hline
\end{tabular}
\end{center}
\end{table}
\begin{figure*}[htbp]
\centerline{\includegraphics[width=\textwidth]{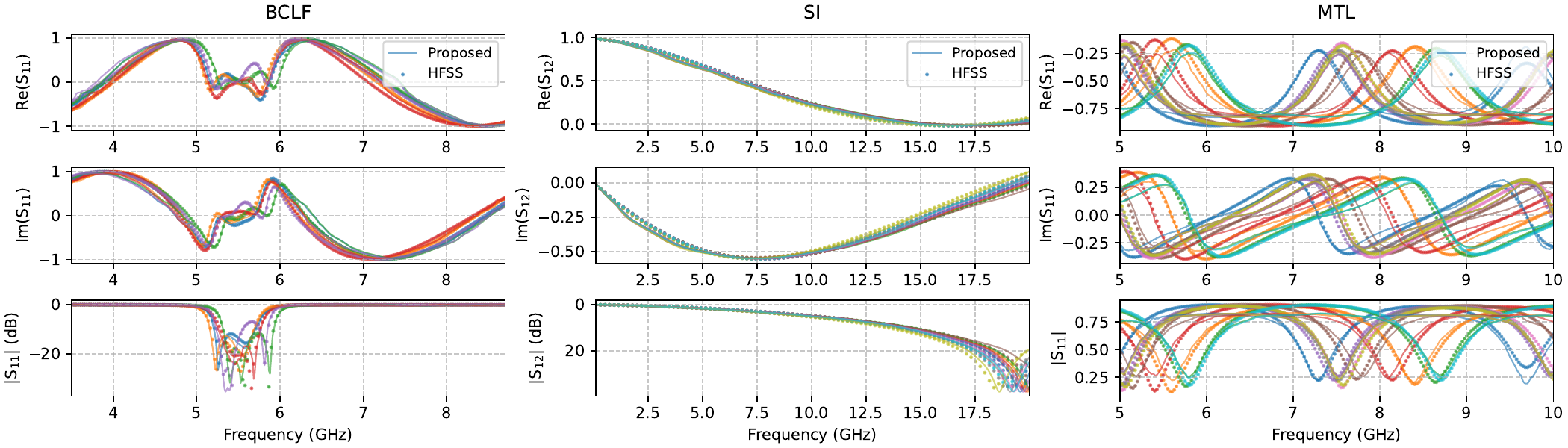}}
\caption{Examples of S-parameters \textbf{estimated} by our proposed BNN models and \textbf{simulation} results generated from Ansys HFSS on three RF components.}
\label{fig:selected}
\end{figure*}
\begin{table*}[htbp]
\centering
\caption{Comparison of conventional ML-based modeling and our proposed Bayesian online learning assisted method on accuracy, sampling method and dataset size}
\label{tab:final}
\begin{tabular}{|l||c|c|c|c||c|c|c|c||c|c|c|c|}
\hline
\hline
Cases & \multicolumn{4}{c||}{BCLF} & \multicolumn{4}{c||}{SI} & \multicolumn{4}{c|}{MTL} \\ \hline
Network & \cite{wang2023mtl} & \multicolumn{3}{c||}{FC-BNN} & \cite{wang2023mtl} & \multicolumn{3}{c||}{FC-BNN} & \cite{torun2019spectral} & \multicolumn{3}{c|}{TC-BNN}\\\hline
Train Size & $1058$ & \multicolumn{3}{c||}{$\mathbf{300}$} & $316$ & \multicolumn{3}{c||}{$\mathbf{180}$} & $1036$ & \multicolumn{3}{c|}{$\mathbf{300}$}\\\hline
Freq. Num. & $401$ & \multicolumn{3}{c||}{$\mathbf{20}$} & $664$ & \multicolumn{3}{c||}{$\mathbf{20}$} & $401$& \multicolumn{3}{c|}{$\mathbf{20}$}\\\hline
Sim. Time (min) & $21.5$ & \multicolumn{3}{c||}{$\mathbf{1.6}$} & $34.0$ & \multicolumn{3}{c||}{$\mathbf{1.5}$} & $16.8$& \multicolumn{3}{c|}{$\mathbf{2.3}$}\\\hline
Geom. Samp. & Full & Random & Random & \textbf{UAW} & Full & Random & Random & \textbf{UAW} & Full & Random & Random & \textbf{UAW}\\ \hline
Freq. Samp. & Full & Uniform & UAW & \textbf{UAW} & Full & Uniform & UAW & \textbf{UAW} & Full & Random & Random & \textbf{UAW} \\ \hline
MSE & $4.98$ & $7.57$ & $7.07$ & $\mathbf{4.85}$ & $1.13$ & $1.99$ & $1.65$ & $\mathbf{0.55}$ & $\mathbf{3.25}$ & $4.76$ & $4.52$ & $\mathbf{3.79}$\\\hline
RMSE & $2.16$ & $2.69$ & $2.60$ & $\mathbf{2.12}$ & $1.03$ & $1.36$ & $1.24$ & $\mathbf{0.70}$ & $\mathbf{1.77}$ & $2.12$ & $2.05$ & $\mathbf{1.91}$\\\hline
R-squared & $0.986$ & $0.979$ & $0.981$ & $\mathbf{0.987}$ & $0.979$ & $0.964$ & $0.970$ & $\mathbf{0.990}$ & $\mathbf{0.983}$ & $0.974$ & $0.976$ & $\mathbf{0.980}$\\\hline
PSNR & $28.10$ & $26.05$ & $26.36$ & $\mathbf{28.30}$ & $30.85$  & $28.49$ & $29.37$ & $\mathbf{34.42}$ & $\mathbf{27.75}$ & $26.25$ & $26.60$ & $\mathbf{27.12}$\\\hline
\hline
\end{tabular}
\end{table*}


\subsection{Comparison with Conventional ML-Based Modeling}
In this section, we present experiment results comparing our proposed Bayesian online learning-assisted methodology with conventional ML-based modeling methodology. For a fair comparison, identical neural network architectures are used in all configurations during the final modeling phase in each case. Following established neural network architectures from prior work \cite{wang2023mtl} and \cite{torun2019spectral}, we implement two baseline models: a fully-connected neural network (FCNN) and a transposed convolutional neural network (TCNN). These deterministic models employ identical layer configurations, kernel sizes, and weight initialization methods as our BNN, differing only in their use of stochastic weights and biases for uncertainty quantification. Model performance is comprehensively evaluated using four metrics: mean squared error (MSE), root mean squared error (RMSE), R-squared coefficient, and peak signal-to-noise ratio (PSNR). We assess multi-port S-parameter modeling accuracy by computing the mean error of absolute S-parameters (in dB) across all possible port combinations. Fig.~\ref{fig:selected} illustrates examples of 1) S-parameters estimated by our proposed BNN models and 2) simulation response generated from Ansys HFSS.

Table~\ref{tab:final} compares results of four settings: 1) conventional ML-based modeling flow, which uses 80\% of geometric design space combinations and full-frequency spectrum simulations to train a deterministic neural network; 2) sparse random sampling with limited geometric combinations and only 20 uniformly distributed frequency points; 3) sparse random sampling on geometry domain combined with uncertainty-aware frequency sampling; and 4) our complete Bayesian online learning method that performs joint adaptive sampling in both domains. 


\textbf{EM Simulation and Training Time.} Table~\ref{tab:final} lists the average EM simulation times per sample, measured on a workstation with an Intel Core i7-9700 CPU and 16GB RAM. The complete BNN training and uncertainty estimation process requires approximately 40 minutes of computational overhead (60,000 epochs in total). This represents a negligible cost compared to the several hours of simulation time saved.Compared with conventional flow, our method requires only 38\% of geometric training points and 13\% of frequency samples, representing only 1.6\% of total EM simulations needed. Our method only requires 2.86\% of total simulation time cost.

\textbf{Inference Accuracy.} The ablation study between setting 2), 3) and 4) demonstrates that the uncertainty-aware sampling method independently reduces modeling error in both the frequency and geometric domains. Our framework outperforms conventional ML-assisted modeling with a 1.9\% RMSE reduction for BCLF and a substantial 32\% improvement for SI structure, while maintaining comparable performance for MTL ($\Delta$RMSE = 7.8\%). 




\section{Conclusion}
This paper presents a Bayesian online learning framework with uncertainty-aware sampling for efficient modeling of RF passive components. The method significantly reduces the need for extensive EM simulations while maintaining high accuracy. Experiment results demonstrate a 35$\times$ speedup compared to conventional ML approaches. The framework offers a promising solution for accelerating RF circuit design with efficient dataset sampling.

\bibliographystyle{IEEEtran}

\bibliography{aspdac_2026}

\end{document}